\documentclass[10pt,twocolumn,letterpaper]{article}

\usepackage{cvpr}              %
\makeatletter
\@namedef{ver@everyshi.sty}{}
\makeatother
\usepackage{tikz}

\usepackage{graphicx}
\usepackage{amsmath}
\usepackage{amssymb}
\usepackage{booktabs}
\usepackage{makecell}
\usepackage{multirow}

\usepackage{amssymb}
\usepackage{pifont}

\newcommand{\poseLoss}[1]{}
\newcommand{\extraAblation}[1]{}
\newcommand{\maxCov}{1.0\xspace}
\newcommand{\meanFlow}{2.5\xspace}

\newcommand{\orange}[1]{{\color{orange}#1}}
\newcommand{\TODO}[1]{
    \orange{{\bf TODO:}~#1} %
}

\newcommand{\nerf}{NeRF\xspace}
\newcommand{\nerfs}{{\nerf}s\xspace}

\usepackage[pagebackref,breaklinks,colorlinks]{hyperref}

\newcommand{\blue}[1]{{\color{blue}#1}}
\newcommand{\linkToPdf}[1]{\href{#1}{\blue{(pdf)}}}
\newcommand{\linkToPpt}[1]{\href{#1}{\blue{(ppt)}}}
\newcommand{\linkToCode}[1]{\href{#1}{\blue{(code)}}}
\newcommand{\linkToWeb}[1]{\href{#1}{\blue{(web)}}}
\newcommand{\linkToVideo}[1]{\href{#1}{\blue{(video)}}}
\newcommand{\linkToMedia}[1]{\href{#1}{\blue{(media)}}}

\usepackage[capitalize]{cleveref}
\crefname{section}{Sec.}{Secs.}
\Crefname{section}{Section}{Sections}
\Crefname{table}{Table}{Tables}
\crefname{table}{Tab.}{Tabs.}

\def\cvprPaperID{*****} %
\def\confName{CVPR}
\def\confYear{2022}

\usepackage{svg}
\usepackage{amsmath}

\begin{document}

\title{NeRF-SLAM: Real-Time Dense Monocular SLAM \\ with Neural Radiance Fields}

\author{
  Antoni Rosinol
  \and
  John J.~Leonard\\
  Massachusetts Institute of Technology \\
  {\tt\small \{arosinol, jleonard, lcarlone\}@mit.edu}
  \and
  Luca Carlone
}

\maketitle

\begin{abstract}
    We propose a novel geometric and photometric 3D mapping pipeline
    for accurate and real-time scene reconstruction from monocular images.
    To achieve this, we leverage recent advances in dense monocular SLAM 
    and real-time hierarchical volumetric neural radiance fields.
    Our insight is that dense monocular SLAM provides the right information to fit 
    a neural radiance field of the scene in real-time,
    by providing accurate pose estimates and depth-maps with associated uncertainty.
    With our proposed uncertainty-based depth loss, 
    we achieve not only good photometric accuracy, but also great geometric accuracy.
    In fact, our proposed pipeline achieves better geometric and photometric accuracy than competing approaches (up to $179\%$ better PSNR and $86\%$ better L1 depth),
    while working in real-time and using only monocular images.
\end{abstract}

\section{Introduction}
\label{sec:intro}

3D reconstruction from monocular images remains one of the most difficult computer vision problems.
Achieving 3D reconstructions in real-time from images alone is a key enabler for applications in robotics,
surveying, and gaming, such as autonomous vehicles, crop monitoring, and augmented reality.

While many 3D reconstruction solutions are based on RGB-D or Lidar sensors,
scene reconstruction from monocular imagery provides a more convenient solution.
RGB-D cameras can fail under certain conditions, such as direct sunlight, and Lidar remains heavier and more expensive than a monocular RGB camera.
Alternatively, stereo cameras simplify the depth estimation problem to a 1D disparity search, but rely on accurate calibration of the cameras that 
is prone to miscalibration during practical operations.
Instead, monocular cameras are inexpensive, lightweight,
and represent the simplest sensor configuration to calibrate.

Unfortunately, monocular 3D reconstruction is a challenging problem due to the lack of explicit measurements of the depth of the scene.
Nonetheless, great progress has been recently made towards monocular 3D reconstructions by leveraging deep-learning approaches.
Given that deep-learning currently achieves the best performance for optical flow \cite{teed2020raft}, and depth \cite{yao2018mvsnet} estimation,
 a plethora of works have tried to use deep-learning modules for SLAM.
For example, using depth estimation networks from monocular images \cite{Tateno17cvpr-CNN-SLAM}, multiple images, as in multi-view stereo \cite{koestler2022tandem},
or using end-to-end neural networks \cite{bloesch2018codeslam}.

\begin{figure*}[h!]
    \centering
    \includegraphics[width=1.0\textwidth,clip, trim={0cm 0cm 0cm 0cm}]{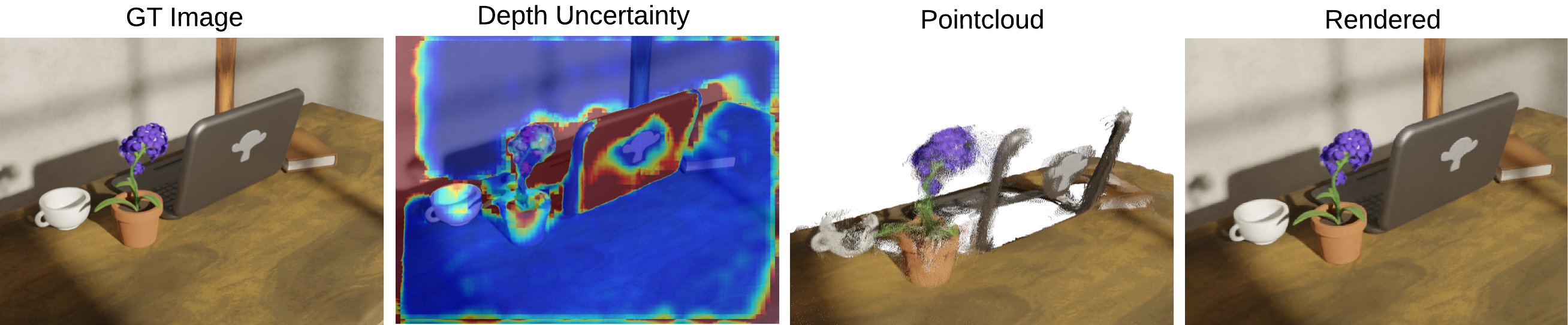}
    \caption{From left to right, input RGB image, estimated depth uncertainty,
      back-projected depth-maps into a pointcloud, after thresholding the depth by its uncertainty ($\sigma_d\leq\maxCov$) for visualization,
      and our resulting neural radiance field, rendered from the same viewpoint as the input image.
      Our pipeline is capable of reconstructing neural radiance fields in real-time given only a stream of RGB images.
    }
    \label{fig:main}
    \vspace{-1em}
\end{figure*}

However, even with the improvements due to deep-learning,
building both geometrically and photometrically accurate 3D maps of the 
scene from a casually taken monocular video in real-time is not currently possible.

Neural Radiance Fields (\nerfs) have recently enabled 3D representations of the world that are photometrically accurate.
Unfortunately, \nerfs are difficult to infer, given the costly volumetric rendering necessary to build a \nerf,
leading to slow reconstructions. Further, \nerfs originally required ground-truth pose estimates to converge.

Nevertheless, recent work shows that it is possible to fit a radiance field in real-time given posed-images \cite{muller2022instant},
while others show that the ground-truth poses are not strictly necessary\cite{lin2021barf,chng2022garf}, as long as a sufficiently good initial estimate is given.
Hence, real-time pose-free \nerf reconstructions could lead to metrically accurate 3D maps.

Despite this, a fundamental problem of \nerf representations,
given no depth supervision, is that the parametrization of the surfaces by a density is prone to `floaters',
ghost geometry that appears because of bad initializations or convergence to bad local minima.
It has been shown that adding depth supervision significantly improves the removal of this ghost geometry, 
and the depth signal leads to faster convergence of radiance fields \cite{deng2022depth}.

Given these findings, our insight is that having a dense monocular SLAM pipeline, that outputs close-to-perfect 
pose estimates, together with dense depth maps and uncertainty estimates, provides the right information
for building neural radiance fields of the scene on the fly.
Our experiments show that this is indeed possible,
and that compared to other approaches we achieve more accurate reconstructions in less time.

\textbf{Contributions} We propose the first scene reconstruction pipeline combining the benefits of
dense monocular SLAM and hierarchical volumetric neural radiance fields.
Our approach builds accurate radiance fields from a stream of images,
without requiring poses or depths as input, and runs in real-time.
We achieve state-of-the-art performance on the Replica dataset for monocular approaches.

\section{Related Work}
\label{sec:related_work}

We review the literature on two different lines of work, dense monocular SLAM and neural radiance fields, and discuss the literature 
at the intersection of both fields.

\subsection{Dense SLAM}

The main challenges to achieve dense SLAM are 
(i) the computational complexity, due to the shear amount of depth variables to be estimated, 
and (ii) dealing with ambiguous or missing information to estimate the depth of the scene, such as textureless surfaces or aliased images.

Historically, the first problem has been bypassed by decoupling the pose and depth estimation.
For example, DTAM \cite{Newcombe2011iccv-dtam} achieves dense SLAM by using the same paradigm as the sparse PTAM\cite{Klein07ismar},
 which tracked the camera pose first and then the depth, in a de-coupled fashion. %
The second problem is also typically avoided by using RGB-D or Lidar sensors, that provide explicit depth measurements, or stereo cameras that simplify depth estimation.

Nevertheless, recent research on dense SLAM has achieved impressive results in these two fronts.
To reduce the number of depth variables,
 CodeSLAM\cite{bloesch2018codeslam} 
optimizes instead the latent variables of an auto-encoder that infers 
depth maps from images. 
By optimizing these latent variables, the dimensionality of the problem is significantly reduced, while the resulting depth maps remain dense.
Tandem\cite{koestler2022tandem} is able to reconstruct 3D scenes 
with only monocular images by using a pre-trained MVSNet-style 
neural-network on monocular depth estimation,
and then decoupling the pose/depth problem by performing frame-to-model photometric tracking.
Droid-SLAM\cite{teed2021droid} shows that by adapting a state-of-the-art dense optical flow estimation architecture \cite{teed2020raft} to the 
problem of visual odometry, it is possible to achieve competitive results in a variety of challenging datasets 
(such as the Euroc \cite{Burri16ijrr-eurocDataset} and TartanAir \cite{wang2020tartanair} datasets),
Droid-SLAM avoids the dimensionality problem by using downsampled depth maps
that are subsequently upsampled using a learned upsampling operator.
Rosinol et al.~\cite{Rosinol22wacv} further show that dense monocular SLAM 
can reconstruct faithful 3D meshes of the scene by weighting the depths estimated in dense SLAM
by their marginal covariance, and subsequently fusing them in a volumetric representation.
The resulting mesh is geometrically accurate,
but due to the limitations of TSDF representations, 
their reconstruction lacks photometric detail and is not fully complete.

Our approach is inspired by the work from Rosinol et al.~\cite{Rosinol22wacv},
where we replace the volumetric TSDF for a hierarchical volumetric neural radiance field as our map representation.
By using radiance fields, %
our approach achieves photometrically accurate maps and improves the completeness of the reconstruction, while also allowing
the optimization of poses and the map simultaneously.

\subsection{Neural Radiance Fields (\nerfs)}

To enable photometrically accurate maps, \nerfs have proved to be a useful representation
that allows capturing view-dependent effects while maintaining multi-view consistency.
\nerf\cite{mildenhall2020nerf} is the seminal work that gave name to this representation, and sparked 
a revolution in terms of new papers that tried to improve several aspects,
the most prominent ones being: reducing the training time to build a \nerf
and the improvement of the underlying 3D reconstruction.

While the vanilla \nerf approach, using one large MLP, requires hours of training to converge,
several authors show that a smaller MLP, combined with 3D spatial data structures to partition the scene, leads to substantial speed-ups.
In particular, NGLOD~\cite{takikawa2021neural} proposes to use tiny MLPs in a volumetric grid,
leading to faster reconstructions, but not quite real-time.
Plenoxels~\cite{yu2021plenoxels} further improved the speed by parametrizing the directional encoding using spherical harmonics, while bypassing the use of an MLP.
Finally, Instant-NGP\cite{muller2022instant} shows that with a hash-based hierarchical volumetric representation of the scene,
it is possible to train a neural radiance field in real-time.

Several works also show how to improve the 3D reconstruction when using radiance fields.
While a variety of works have proposed to replace the density field by a signed distance functions or other representations\cite{oechsle2021unisurf, yu2022monosdf, wang2021neus, yariv2021volume},
we are most interested on the insights related to the use of alternative supervisory signals to 
increase the convergence speed of NeRFs, as well as the quality of the underlying geometry~\cite{wang2022neuris,yu2022monosdf}.
In particular, Mono-SDF~\cite{yu2022monosdf} shows that state-of-the-art deep learning models for depth and normal estimation from monocular images
provides useful information that can significantly improve the convergence speed and quality of the radiance field reconstruction. %

Our work leverages these insights by using the information provided from dense SLAM, which estimates poses and dense depth maps.
We also use the fact that dense SLAM outputs are probabilistic in nature, and use this information,
which is typically discarded in current approaches, to weight the supervisory signals to fit a radiance field.

\subsection{SLAM with \nerfs}

Another important axis of research in neural radiance fields 
is to remove its dependency on partially known camera poses.
This is particularly enticing for building \nerfs without having to process the data to get the camera poses of the images,
a task that is usually long and is typically done using COLMAP\cite{Schonberger16cvpr-SfMRevisited}.

iNeRF~\cite{yen2021inerf} was the first to show that it is possible to regress the camera pose given an already built \nerf of the scene.
It also made evident that the basin of convergence for pose estimation is small, as it is typical from direct image alignment.
Barf~\cite{lin2021barf} further shows how to simultaneously fit a \nerf while estimating camera poses, given an inaccurate initial guess, by 
formulating the optimization as an iterative image alignment problem.
They also propose to modulate \nerf's positional encoding (similarly to FourierFeatures\cite{tancik2020fourier}) to imitate 
coarse-to-fine approaches that improve the basin of convergence for direct approaches.
Unfortunately, these approaches are too slow for online inference due to their choice of a large MLP as map representation.

iMap\cite{sucar2021imap} and Nice-SLAM\cite{zhu2022nice} 
subsequently showed how to build accurate 3D reconstructions, without the need for poses, by partially de-coupling pose and depth estimation, 
similarly to our approach, and by using RGB-D images. %
While iMAP used a single MLP for the whole scene, Nice-SLAM leveraged the learnings on volumetrically partitioning the space,
 as detailed above, to make inference faster and more accurate.
Both approaches use depth from RGB-D cameras as input,
instead, our work only uses monocular images.

Finally, VolBA~\cite{clark2022volumetric}, Orbeez-SLAM~\cite{chung2022orbeez} and Abou-Chakra et al.~\cite{abou2022implicit}
also use a hierarchical volumetric map for real-time SLAM, but this time from monocular images.
VolBA uses a direct RGB loss for pose estimation and mapping,
while Orbeez-SLAM and Abou-Chakra et al. use ORB-SLAM~\cite{Mur-Artal17tro-ORBSLAM2} to provide the initial poses.
By using dense monocular SLAM, our approach both leverages an indirect loss for pose estimation,
which is known to be more robust than direct image alignment, %
and has the ability to depth-supervise the radiance field,
which is known to improve quality and convergence speed. %
Further, differently than other depth-supervised approaches \cite{deng2022depth,yen2022nerf,sucar2021imap,dey2022mip,azinovic2022neural},
our depth loss is weighted by the depth's marginal covariance, 
and the depth is estimated from optical-flow rather than measured by an RGB-D camera.

Overall, our work leverages recent work on dense monocular SLAM (Droid-SLAM\cite{teed2021droid}),
probabilistic volumetric fusion (Rosinol et al.~\cite{Rosinol22wacv}),
and hash-based hierarchical volumetric radiance fields (Instant-NGP~\cite{muller2022instant}),
to estimate geometric and photometric maps of the scene in real-time, without the need of depth images or poses.

\section{Methodology}

\begin{figure}[h!]
    \centering
    \includegraphics[width=1.0\columnwidth, clip, trim={0cm 0cm 0cm 0cm}]{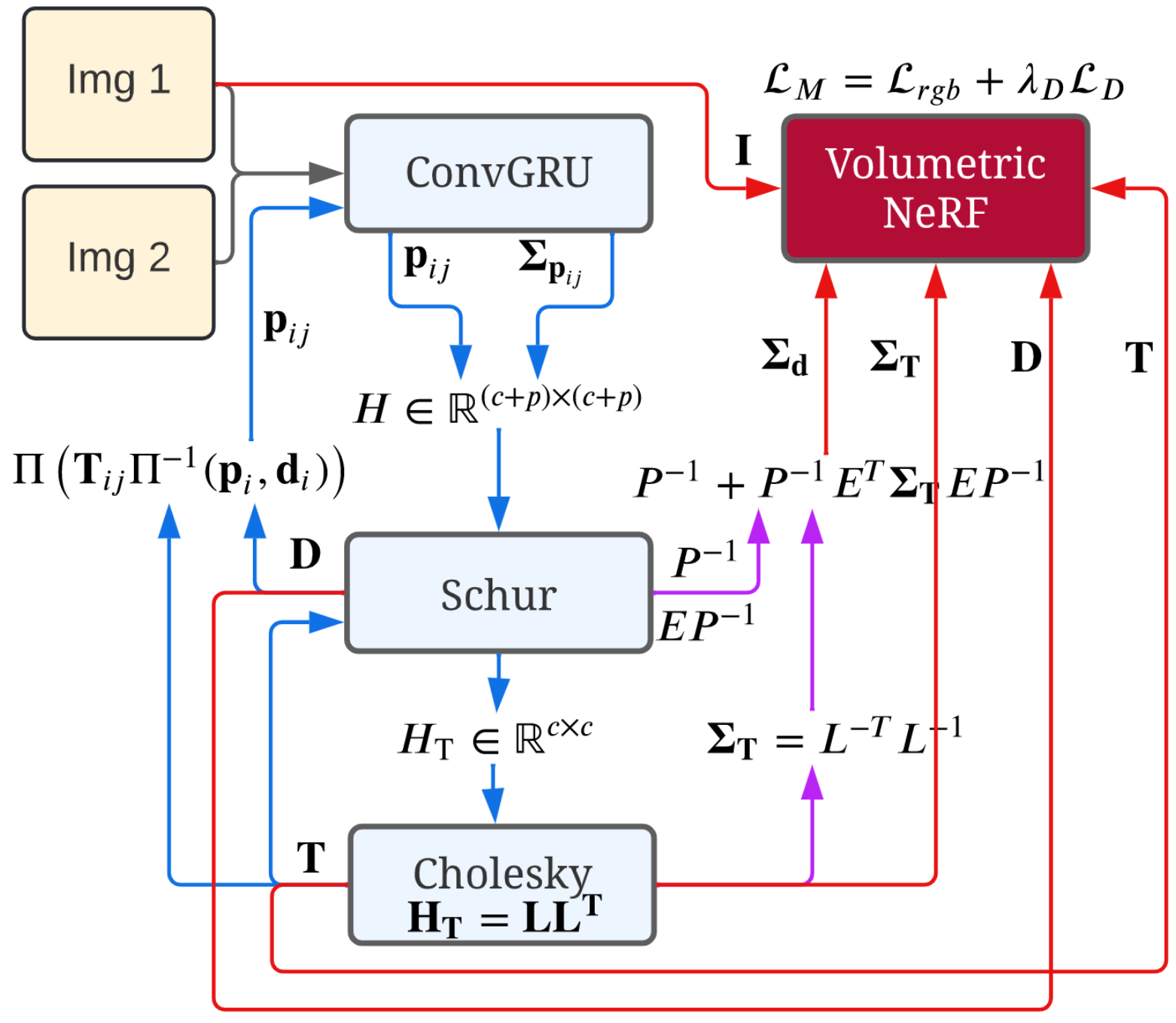}
    \caption{The input to our pipeline consists of sequential monocular images (here represented as Img 1 \& Img 2).
    Starting from the top-right, our architecture fits a \nerf using Instant-NGP~\cite{muller2022instant}, 
    which we supervise using RGB images $\mathbf{I}$, depths $\mathbf{D}$,\poseLoss{ and poses $\mathbf{T}$,}
    where the depths \poseLoss{and pose supervision} are weighted by their marginal covariance,
    $\mathbf{\Sigma_D}$\poseLoss{ and $\mathbf{\Sigma_T}$, respectively}.
    Inspired by Rosinol et al.~\cite{Rosinol22wacv}, we compute these covariances from dense monocular SLAM.
    In our case, we use Droid-SLAM~\cite{teed2021droid}.
    We provide more details about the flow of information in \cref{sec:tracking}.
    In blue, we show Droid-SLAM's~\cite{teed2021droid} contributions and flow of information, 
    similarly, in pink are Rosinol's contribution~\cite{Rosinol22wacv}, and in red, our contribution.
    }
    \vspace{-1em}
    \label{fig:architecture}
\end{figure}

The main idea of our approach is to supervise a neural radiance field using the output from dense monocular SLAM.
Dense monocular SLAM can estimate dense depth maps and camera poses, while also providing uncertainty estimates for both depths and poses.
With this information, we can train a radiance field with a dense depth loss weighted by the depths' marginal covariances\poseLoss{, as well as provide uncertainty-weighted camera pose priors}.
By using real-time implementations of both dense SLAM and radiance field training, and by running these in parallel, we achieve real-time performance.
\Cref{fig:architecture} shows the flow of information in our pipeline.
We now explain our architecture, starting with our tracking frontend (\cref{sec:tracking}) and following with our mapping backend (\cref{sec:mapping}).

\subsection{Tracking: Dense SLAM with Covariances}
\label{sec:tracking}

We use as our tracking module Droid-SLAM~\cite{teed2021droid}, which provides dense depth maps and poses for every keyframe.
Starting from a sequence of images,
Droid-SLAM first computes the dense optical-flow $\mathbf{p}_{ij}$ between pairs of frames $i$ and $j$,
using a similar architecture to Raft~\cite{teed2020raft}.
At the core of Raft is a Convolutional GRU (ConvGRU in \cref{fig:architecture}) that,
given the correlation between pairs of frames and a guess of the current optical flow $\mathbf{p}_{ij}$,
computes a new flow $\mathbf{p}_{ij}$, as well as a weight $\mathbf{\Sigma}_{\mathbf{p}_{ij}}$ for each optical flow measurement.

With these flows and weights as measurements, Droid-SLAM solves a dense bundle adjustment (BA) problem
where the 3D geometry is parametrized as a set of inverse depth maps per keyframe.
This parametrization of the structure leads to an extremely efficient way of solving the dense BA problem,
which can be formulated as a linear least-squares problem
by linearizing the system of equations into the familiar cameras/depths arrow-like block-sparse Hessian $H\in \mathbb{R}^{(c+p)\times(c+p)}$,
where $c$ and $p$ are the dimensionality of the cameras and the points.

As can be seen in \cref{fig:architecture}, to solve the linear least-squares problem,
we take the Schur complement of the Hessian to compute the reduced camera matrix $H_T$, which does not depend on the depths, and has a much smaller dimensionality of $\mathbb{R}^{c\times c}.$
The resulting smaller problem over the camera poses is solved by taking the Cholesky factorization of $H_T = LL^T$,
where $L$ is the lower-triangular Cholesky factor, and then solving for the poses $T$ by front and back-substitution.
As shown at the bottom of \cref{fig:architecture}, given these poses $T$, we can solve for the depths $d$.
Furthermore, given poses $T$ and depths $D$,
Droid-SLAM proposes to compute the induced optical-flow and feeds it again as an initial guess to the ConvGRU network,
as seen on the left side of \cref{fig:architecture}, where $\Pi$ and $\Pi^{-1}$, are the projection and back-projection functions.
The blue arrows in \cref{fig:architecture} show the tracking loop, and corresponds to Droid-SLAM.

Then, inspired by Rosinol et al.~\cite{Rosinol22wacv}, we further compute the marginal covariances of both the dense depth maps 
and the poses from Droid-SLAM (purple arrows in \cref{fig:architecture}).
For this, we need to leverage the structure of the Hessian, which we block-partition as follows:
\begin{equation}
    \label{eq:hessian}
    H\mathbf{x} = \mathbf{b}, \quad \text{\ie} \quad 
    \left[
        \begin{array}{cc}
            C & E \\
            E^{T} & P
        \end{array}
    \right]
    \left[
        \begin{array}{l}
            \Delta \boldsymbol{\xi} \\
            \Delta \mathbf{d}
        \end{array}
    \right]
    =
    \left[
        \begin{array}{c}
            \mathbf{v} \\
            \mathbf{w}
        \end{array}
    \right],
\end{equation}
where $H$ is the Hessian matrix, $\mathbf{b}$ the residuals, $C$ is the block camera matrix,
and $P$ is the diagonal matrix corresponding to the inverse depths per pixel per keyframe.
We represent by $\Delta\boldsymbol{\xi}$ the delta updates on the lie algebra of the camera poses in $SE(3)$,
while $\Delta\mathbf{d}$ is the delta update to the per-pixel inverse depths.
$E$ is the camera/depth off-diagonal Hessian's block matrices, and $v$ and $w$ correspond to the pose and depths residuals.

From this block-partitioning of the Hessian, we can efficiently calculate the marginal covariances for the dense depths $\mathbf{\Sigma_d}$ and poses $\mathbf{\Sigma_T}$, as shown in \cite{Rosinol22wacv}:
\begin{equation}
    \label{eq:cov}
    \begin{aligned}
        \mathbf{\Sigma}_{\mathbf{d}} &= P^{-1}+P^{-T} E^T\mathbf{\Sigma}_{\mathbf{T}} E P^{-1} \\
        \mathbf{\Sigma}_{\mathbf{T}} &= (LL^T)^{-1}.\\
    \end{aligned}
\end{equation}
We refer to \cite{Rosinol22wacv} for details on how to compute these in real-time.
    
Finally, given all the information computed by the tracking module -- the poses, the depths, their respective marginal covariances, as well as the input RGB images -- 
we can optimize our radiance field's parameters and refine the camera poses simultaneously.

\subsection{Mapping: Probabilistic Volumetric NeRF}
\label{sec:mapping}

Given the dense depth-maps for each keyframe, it is possible to depth-supervise our neural volume.
Unfortunately, the depth-maps are extremely noisy due to their density,
since even textureless regions are given a depth value.
\Cref{fig:replica_all} shows that the resulting pointcloud from dense monocular SLAM is particularly noisy and contains large outliers (top image in \cref{fig:replica_all}).
Supervising our radiance field given these depth-maps can lead to biased reconstructions, as shown later in \cref{sec:ablations}.

Rosinol et al.\cite{Rosinol22wacv} shows that the uncertainty of the depth estimates is an excellent signal to weight the depth values for classical TSDF volumetric fusion.
Inspired by these results, we use the depth uncertainty estimation to weight the depth loss which we use to supervise our neural volume.
\Cref{fig:main} shows the input RGB image, its corresponding depth-map uncertainty,
the resulting pointcloud (after thresholding its uncertainty by $\sigma_d \leq \maxCov$ for visualization),
and our results when using our uncertainty-weighted depth loss.

\poseLoss{Since we also have pose uncertainties, we can use pose priors, similarly weighted by their marginal covariance, to guide the optimization.
Using pose priors avoids camera poses from drifting or diverging, which may happen if poses are just given as an initial guess.
Since Rosinol et al.\cite{Rosinol22wacv}'s map representation (volumetric TSDF) is not differentiable with respect to the camera poses,
they could not show how poses, weighted by their uncertainty, could be used to simultaneously optimize the poses and the map's parameters.}

Given the uncertainty-aware losses, we formulate our mapping loss as:
\begin{equation}
    \label{eq:mapping_loss}
    \mathcal{L}_M\left(\mathbf{T}, \Theta\right) = \mathcal{L}_{\text{rgb}}\left(\mathbf{T}, \Theta\right) + \lambda_D\mathcal{L}_{\text{D}}\left(\mathbf{T}, \Theta\right) \poseLoss{+ \lambda_T\mathcal{L}_{\text{T}}\left(\mathbf{T}\right)}
\end{equation}
which we minimize with respect to both poses $\mathbf{T}$ and neural parameters $\Theta$, given hyper-parameter\poseLoss{s} $\lambda_D\poseLoss{, \lambda_T}$ balancing\poseLoss{ pose,} depth and color supervision (we set $\lambda_D$ to $1.0$).
In particular, our depth loss is given by:
\begin{equation}
    \mathcal{L}_{\text{D}}\left(\mathbf{T}, \Theta\right) = \|D - D^\star(\mathbf{T}, \Theta)\|^2_{\Sigma_D},
\end{equation}
where $D^\star$ is the rendered depth, and $D, \Sigma_D$ are the dense depth and uncertainty as estimated by the tracking module.
We render the depth $D^\star$ as the expected ray termination distance, similarly to \cite{mildenhall2020nerf, deng2022depth}.
Each depth per pixel is computed by sampling 3D positions along the pixel's ray, evaluating the density $\sigma_i$ at sample $i$,
and alpha-compositing the resulting densities, similarly to standard volumetric rendering:
\begin{equation}
    \begin{aligned}
        d^\star &= \sum_{i} \mathcal{T}_i\Bigl(1-\exp \left(-\sigma_i\delta_{i}\right)\Bigl) d_i,
    \end{aligned}
\end{equation}
with $d_i$ the depth of a sample $i$ along the ray, and $\delta_i = d_{i+1} - d_i$ the distance between consecutive samples.
$\sigma_i$ is the volume density, generated by evaluating an MLP at the 3D world coordinate of sample $i$.
We refer to \cite{muller2022instant} for more details on the inputs given to the MLP.
Lastly, $\mathcal{T}_i$ is the accumulated transmittance along the ray up to sample $i$, defined as:
\begin{equation}
    \label{eq:transmittance}
    \mathcal{T}_i = \exp \Bigl(-\sum_{j<i} \sigma_{j}\delta_j\Bigl).
\end{equation}
Our color loss is defined as in the original \nerf~\cite{mildenhall2020nerf}:
\begin{equation}
    \mathcal{L}_{\text{rgb}}\left(\mathbf{T}, \Theta\right) = \|I - I^\star(\mathbf{T}, \Theta)\|^2,
\end{equation}
where $I^\star$ is the rendered color image, synthesized similarly to the depth image, by using volumetric rendering.
Each color per pixel is likewise computed by sampling along the pixel's ray and alpha-compositing the resulting densities and colors:
$\sum_i \mathcal{T}_i\bigl(1-\exp \left(-\sigma_i \delta_i\right)\bigl) \mathbf{c}_i,$
where $\mathcal{T}_i$ is again the transmittance as in \cref{eq:transmittance},
and $\mathbf{c}_i$ is the color estimated by the MLP.
Both the density $\sigma_i$ and the color $\mathbf{c}_i$ are estimated simultaneously for a given sample $i$. %

\poseLoss{Finally, our pose loss is a covariance-weighted prior, \ie:
\begin{equation}
    \mathcal{L}_{\text{T}}\left(\mathbf{T}\right) = \|\mathbf{T} - \mathbf{T}^\star\|^2_\mathbf{\Sigma_T}.
\end{equation}
where $\mathbf{T}^\star$ is the measured pose estimate by our tracking frontend, with $\mathbf{\Sigma_T}$ the marginal covariance of the pose.
}

The mapping thread continuously minimizes our mapping loss function $\mathcal{L}_M(\mathbf{T}, \Theta)$ (\cref{eq:mapping_loss}).
In \cref{sec:results}, we show that this approach achieves accurate results while running in real-time.

\subsection{Architecture}
\label{ssec:architecture}

Our pipeline consists of a tracking and a mapping thread, both running in real-time and in parallel.
The tracking thread continuously minimizes the BA re-projection error for an active window of keyframes.
The mapping thread always optimizes all of the keyframes received from the tracking thread,
and does not have a sliding window of active frames.

The only communication between these threads happens when the tracking pipeline generates a new keyframe.
On every new keyframe, the tracking thread sends the current keyframes' poses with their respective images and estimated depth-maps,
 as well as the \poseLoss{pose and} depths' marginal covariances, to the mapping thread.
Only the information currently available in the sliding optimization window of the tracking thread is sent to the mapping thread.
The active sliding window of the tracking thread consists of at most $8$ keyframes.
The tracking thread generates a new keyframe as soon as the mean optical-flow between the previous keyframe and 
the current frame is higher than a threshold (in our case $\meanFlow$ pixels).

Finally, the mapping thread is also in charge of rendering for interactive visualization of the reconstruction.

\subsection{Implementation Details}
\label{ssec:details}

We perform all computations in Pytorch and CUDA,
and use an RTX 2080 Ti GPU for all our experiments (11Gb of memory).
We use Instant-NGP\cite{muller2022instant} as our hierarchical volumetric neural radiance field map,
which we modify to add our proposed mapping loss $\mathcal{L}_M$ in \cref{eq:mapping_loss}.
As our tracking frontend, we use Droid-SLAM\cite{teed2021droid}, and re-use their pre-trained weights.
We compute depth and pose uncertainties using Rosinol's approach~\cite{Rosinol22wacv} which computes uncertainties in real-time.
We use the same GPU for both tracking and mapping, although our approach allows the use of two separate GPUs for tracking and mapping.

\section{Results}
\label{sec:results}

\begin{table*}[h!]
  \centering
  \footnotesize
  \setlength{\tabcolsep}{0.36em}
  \renewcommand{\arraystretch}{1.5}
    \begin{tabular}{clcccccccccccccccccc}
      \toprule
         & & \multicolumn{1}{c}{\makecell{\tt{room-0}}} & \multicolumn{1}{c}{\makecell{\tt{room-1}}} &  \multicolumn{1}{c}{\makecell{\tt{room-2}}} & \multicolumn{1}{c}{\makecell{\tt{office-0}}} & \multicolumn{1}{c}{\makecell{\tt{office-1}}} & \multicolumn{1}{c}{\makecell{\tt{office-2}}}& \multicolumn{1}{c}{\makecell{\tt{office-3}}} & \multicolumn{1}{c}{\makecell{\tt{office-4}}} & Avg. \\
        \midrule
            \multirow{2}{*}{\makecell{\textbf{iMAP$^*$}~\cite{sucar2021imap}\\(GT depth)}}
              & {\bf Depth L1} [cm] $\downarrow$
              & 5.70 & 4.93 & 6.94 & 6.43 & 7.41 & 14.23 & 8.68 & 6.80 & 7.64\\
              & {\bf PSNR } [dB] $\uparrow$
              & 5.66 & 5.31 & 5.64 & 7.39 & 11.89 & 8.12 & 5.62 & 5.98 & 6.95\\
        \midrule
            \multirow{2}{*}{{\makecell{\textbf{Nice-SLAM}~\cite{zhu2022nice} \\  (GT depth)}}}
            & {\bf Depth L1} [cm] $\downarrow$
            & \textbf{ 2.53 } & \textbf{ 3.45 } & \textbf{ 2.93 } & \textbf{ 1.51 } & \textbf{ 0.93 } & \textbf{ 8.41 } & \textbf{ 10.48 } & \textbf{ 2.43 } & \textbf{ 4.08 } \\
            & {\bf PSNR } [dB] $\uparrow$
            & \textbf{ 29.90 } & \textbf{ 29.12 } & \textbf{ 19.80 } & \textbf{ 22.44 } & \textbf{ 25.22 } & \textbf{ 22.79 } & \textbf{ 22.94 } & \textbf{ 24.72 } & \textbf{ 24.61 } \\
        \midrule
        \midrule
            \multirow{2}{*}{\makecell{\textbf{TSDF-Fusion} \\ Res. = 256 \\ (our depth)}}
            & {\bf Depth L1} [cm] $\downarrow$
            & 23.51 & 20.94 & 23.34 & 14.11 & 10.50 & 30.89 & 28.92 & 22.83 & 21.88 \\
            & {\bf PSNR } [dB] $\uparrow$
            & 3.43 & 4.51 & 5.57 & 11.16 & 15.92 & 4.86 & 5.68 & 5.46 & 7.07 \\

        \midrule
            \multirow{2}{*}{\makecell{\textbf{$\sigma$-Fusion}\cite{Rosinol22wacv} \\ Res. = 256 \\ (our depth)}}
            & {\bf Depth L1} [cm] $\downarrow$
            & 21.92 & 19.28 & 22.40 & 13.80 & 10.21 & 22.27 & 28.70 & 22.21 & 20.10 \\
            & {\bf PSNR } [dB] $\uparrow$
            & 3.45  & 4.51 & 5.57 & 11.16 & 15.92 & 4.86 & 5.69 & 5.46 & 7.08 \\

        \midrule
            \multirow{2}{*}{{\makecell{\textbf{Nice-SLAM}~\cite{zhu2022nice}\\ (no depth)}}}
            & {\bf Depth L1} [cm] $\downarrow$
            & 11.12 & 9.42 & 19.03 & 11.12 & 10.24 & 16.36 & 21.33 & 14.81 & 14.18 \\
            & {\bf PSNR } [dB] $\uparrow$
            & 18.15 & 18.22 & 17.82 & 20.23 & 19.14 & 15.22 & 16.12& 17.24  & 17.76 \\

        \midrule
            \multirow{2}{*}{{\makecell{\textbf{Ours} \\ (our depth) }}}
            & {\bf Depth L1} [cm] $\downarrow$
            & \textbf{2.97}  & \textbf{2.63}  & \textbf{2.58}  & \textbf{2.49}   & \textbf{1.98}  & \textbf{9.13}  & \textbf{10.58} & \textbf{3.59} & \textbf{4.49} \\
            & {\bf PSNR } [dB] $\uparrow$
            & \textbf{34.90} & \textbf{36.95} & \textbf{40.75} & \textbf{48.07}  & \textbf{53.44} & \textbf{39.30} & \textbf{38.63} & \textbf{39.21} & \textbf{41.40} \\
      \bottomrule
    \end{tabular}%
    \caption{Geometric (L1) and Photometric (PSNR) results for the Replica dataset. iMAP and Nice-SLAM are first evaluated using the ground-truth depth from Replica as supervision (top two rows).
    We also evaluate Nice-SLAM when not using ground-truth depth as supervision for comparison.
    TSDF-Fusion, $\sigma$-Fusion, and our approach are evaluated using the poses and depths from dense monocular SLAM, as explained in \cref{sec:tracking}.
    Best results in bold.}
    \label{tab:replica_per_scene}
\end{table*}

In the following, we evaluate our approach against competing techniques,
and provide ablation experiments to evaluate the improvements deriving from our proposed architecture and suggested uncertainty-aware loss.

\subsection{Datasets}

We use two datasets for evaluation: the Cube-Diorama dataset~\cite{abou2022implicit} and the Replica dataset~\cite{straub2019replica}.

Cube-Diorama is a synthetic dataset generated with Blender, which provides ground-truth poses, depths, and images.
Using this dataset, we are able to do accurate ablation experiments to evaluate the benefits of our proposed approach.

The Replica dataset provides real-world scenes which we use to evaluate and compare our approach with related work.
The dataset consists of high quality 3D reconstructions of 5 offices and 3 apartments. 
For evaluation, we use the data generated by rendering a random trajectory of 2000 RGB and depth frames per scene (as generated by the authors of iMAP\cite{sucar2021imap}).
The rendered depth-maps can be considered ground-truth, since they have been rendered from the ground-truth mesh.

\subsection{Methods for Evaluation}

The approaches we are comparing are classical TSDF-fusion~\cite{Curless96siggraph}, 
probabilistic TSDF-Fusion from Rosinol et al.~\cite{Rosinol22wacv} (which we label as $\sigma$-Fusion),
 iMAP~\cite{sucar2021imap}, and Nice-SLAM~\cite{zhu2022nice}.
This allows us to compare geometric and probabilistic approaches (TSDF-Fusion, $\sigma$-Fusion),
as well as learning-based approaches (iMAP and Nice-SLAM).

TSDF-Fusion and $\sigma$-Fusion use a hash-based volumetric representation to fuse posed depth-maps estimated from our tracking module.
While TSDF-Fusion uses a uniform weight for the depth-maps, $\sigma$-Fusion uses the depth-maps uncertainties to weight down the depths, similarly to ours.
iMAP uses one large MLP to represent the 3D scene, while Nice-SLAM uses hierarchical dense volumetric grids for mapping, similarly to our choice of hash-based hierarchical volumes.

Finally, both iMAP and Nice-SLAM use the depth-maps rendered from the ground-truth meshes
as measurements, which leads to better results than what we would reasonably expect from a casually moving RGB-D camera.
Therefore, we also evaluate Nice-SLAM without ground-truth depth-maps as supervisory signal.

\begin{figure}[h!]
    \centering
    \includegraphics[width=1.0\columnwidth, clip, trim={0cm 0cm 0cm 0cm}]{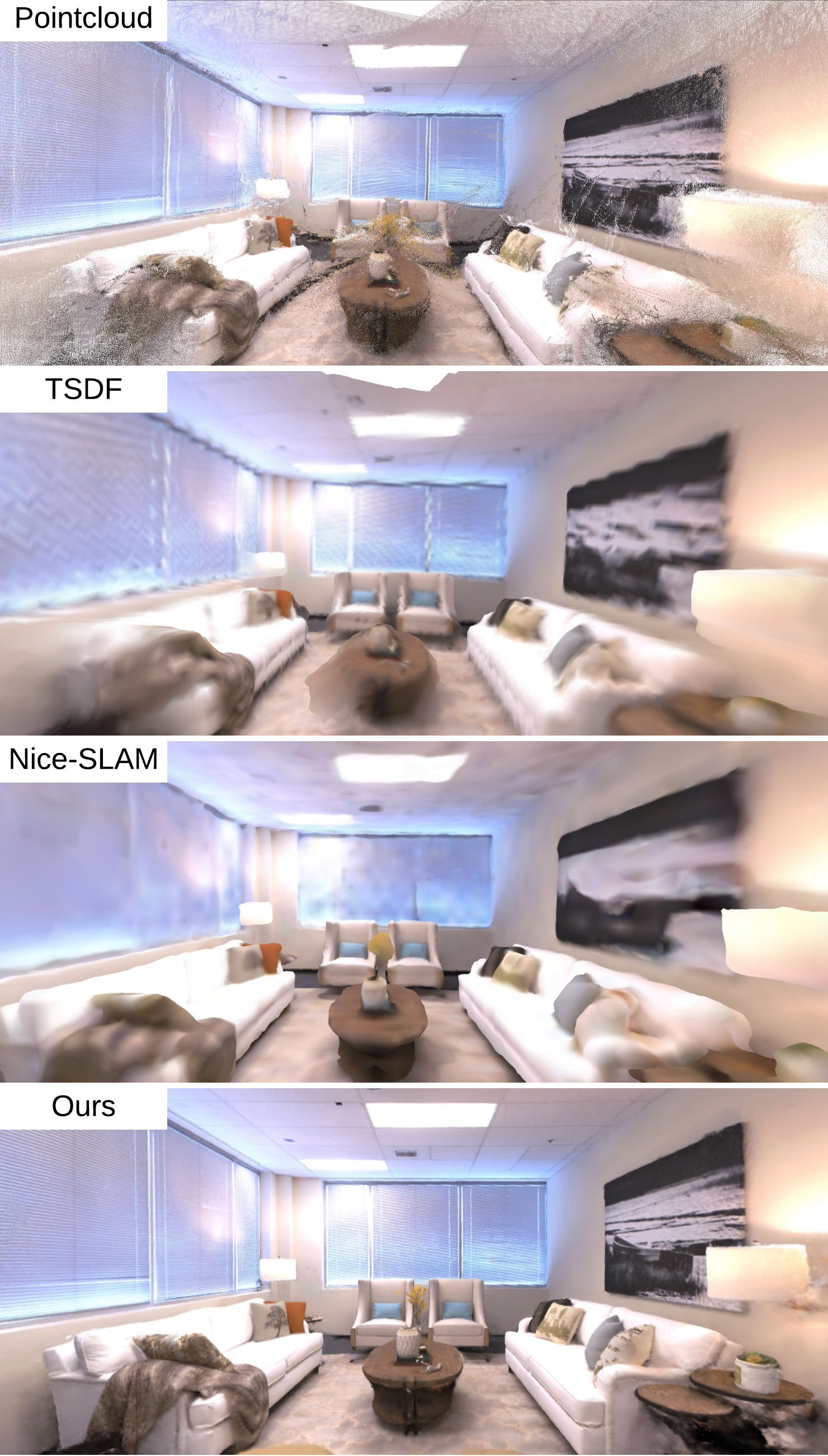}
    \caption{Qualitative results on the Replica {\tt office-0} dataset using different mapping approaches.
     From top to bottom, raw pointcloud from our tracking module,
     TSDF reconstruction using $\sigma$-Fusion, Nice-SLAM's results, and ours.}
    \label{fig:replica_all}
    \vspace{-1em}
\end{figure}

\subsection{Geometric and Photometric Accuracy}

The Replica dataset allows us to evaluate the geometric and photometric quality of the different approaches.
In particular, we use the L1 depth error between the estimated and the ground-truth depth-maps as a proxy for geometric accuracy 
(Depth L1), as well as the peak signal-to-noise ratio (PSNR) between the input RGB images and the rendered images for photometric accuracy.

\Cref{tab:replica_per_scene} shows the results achieved by the different approaches we evaluate.
The first two rows correspond to iMAP and Nice-SLAM under their default setup, with ground-truth depth used as supervisory signal.
Nice-SLAM is superior to iMAP in terms of geometric and photometric accuracy.
The following rows correspond to approaches that do not use the ground-truth depth-maps.
We run TSDF-Fusion and $\sigma$-Fusion with the poses and depth-maps estimated by our tracking module,
 with $\sigma$-Fusion further using the depths' uncertainties.
While $\sigma$-Fusion achieves better geometric reconstructions than TSDF-Fusion, both achieve poor photometric accuracy (we render the resulting 3D mesh extracted using marching cubes).
We also run Nice-SLAM without ground-truth depth.
As expected, Nice-SLAM's results deteriorate when not using ground-truth depth, but it still achieves competitive results when compared to TSDF fusion, particularly in terms of photometric accuracy (PSNR).
Finally, our approach achieves the best results in terms of photometric accuracy,
and geometric accuracy by a substantial margin in most scenes.
Compared to Nice-SLAM, we achieve up to $179\%$ PSNR improvements in office-1, and up to $86\%$ better L1 accuracy in room-2, with the best combined improvements seen in office-1 ($179\%$ better PSNR, $80\%$ better L1).

\Cref{fig:replica_all} provides a qualitative comparison between the different map representations and approaches, where we render the different 3D representations for comparison.
The top image is the raw pointcloud estimated by our tracking module,
where each 3D point is generated by back-projecting the pixel's depth and colored using the pixel's RGB values.
The second image corresponds to the rendered 3D mesh extracted using marching cubes from the volumetric TSDF map built using $\sigma$-Fusion.
The next two images are rendered images from Nice-SLAM and our approach.

Overall, our approach achieves noticeably better results than the other representations.
Remarkably, while the TSDF representation (using $\sigma$-Fusion) uses the same pose, depth, and uncertainty estimates
than our approach, it achieves significantly worst results compared to our approach.

\begin{figure}[h!]
    \centering
    \includegraphics[width=1.0\columnwidth, clip, trim={0cm 0cm 0cm 0cm}]{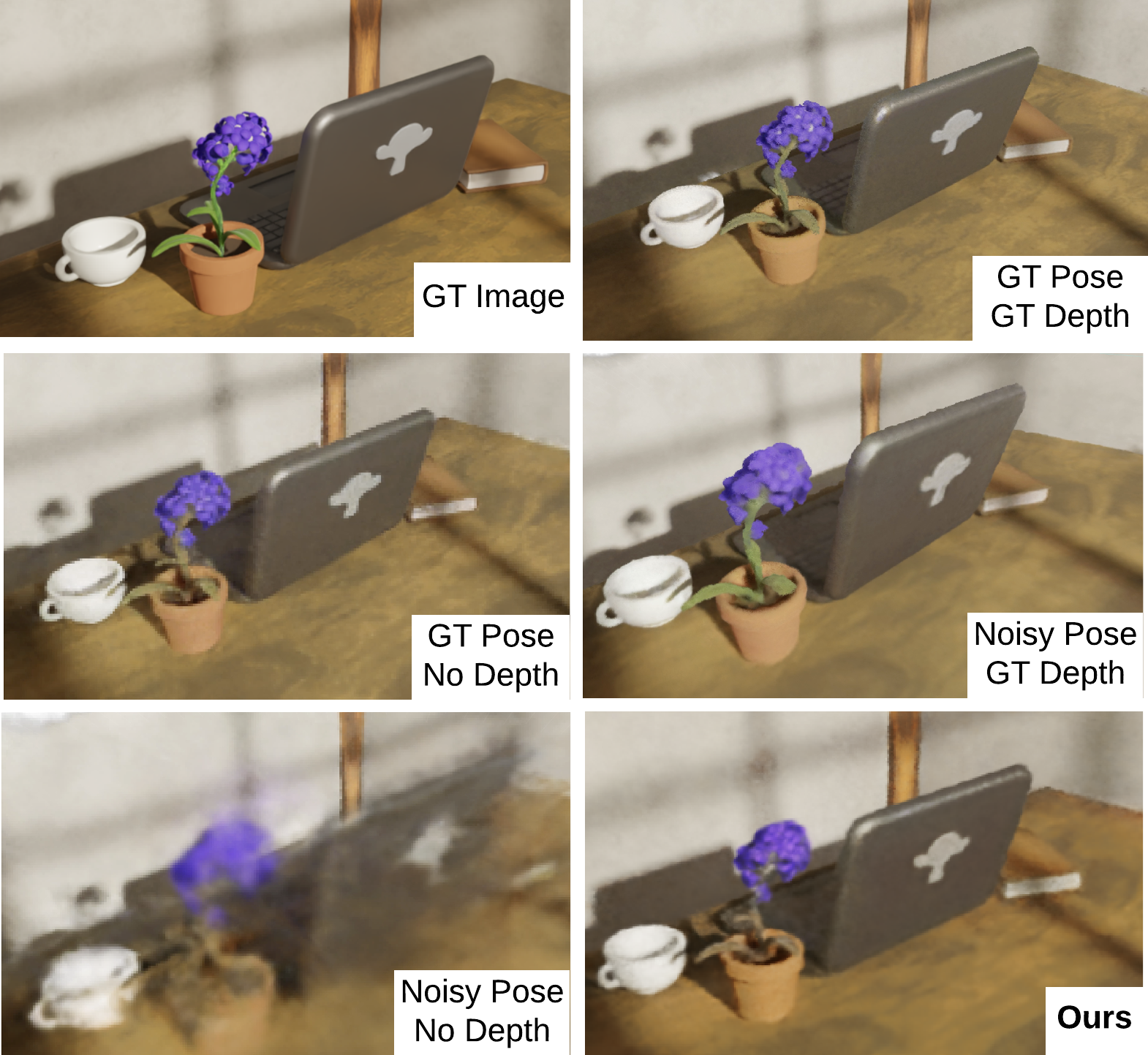}
    \caption{Impact on the performance when using depth supervision with and without ground-truth depth, and 
    when initializing the poses with ground-truth or noisy poses; compared with our approach which estimates dense depths and poses.
    Results after $60s$ of convergence.
    }
    \vspace{-1em}
    \label{fig:proof_of_concept}
\end{figure}

\subsection{Depth Loss Ablation}
\label{sec:ablations}

Depth supervision of neural radiance fields from raw depth-maps,
either estimated from dense SLAM or coming from RGB-D,
is prone to errors, because depth-maps are most often noisy and with outliers.
For dense monocular SLAM, this is particularly problematic, since depth values are estimated even for textureless or aliased regions.

\cref{fig:proof_of_concept} shows that the ideal scenario is to use ground-truth pose and ground-truth depth 
for fast and accurate neural radiance field reconstruction (top-right image).
If ground-truth depths are not provided, but ground-truth poses are available, the radiance field also converges, although at a slower pace (mid-left); this is the classical input for \nerf (posed images).
Instead, if we provide noisy poses and no depth-maps, the radiance field does not converge in less than $60s$ (bottom-left),
while using the ground-truth depth and noisy poses still leads to great results (mid-right).
Our approach aims to reach this last result. 
The bottom-right image in \cref{fig:proof_of_concept} shows that our approach can achieve great results despite using noisy poses and depths,
as long as these are weighted by their uncertainty.

\begin{figure}[h!]
    \centering
    \includegraphics[width=1.0\columnwidth, clip, trim={0cm 0cm 0cm 0cm}]{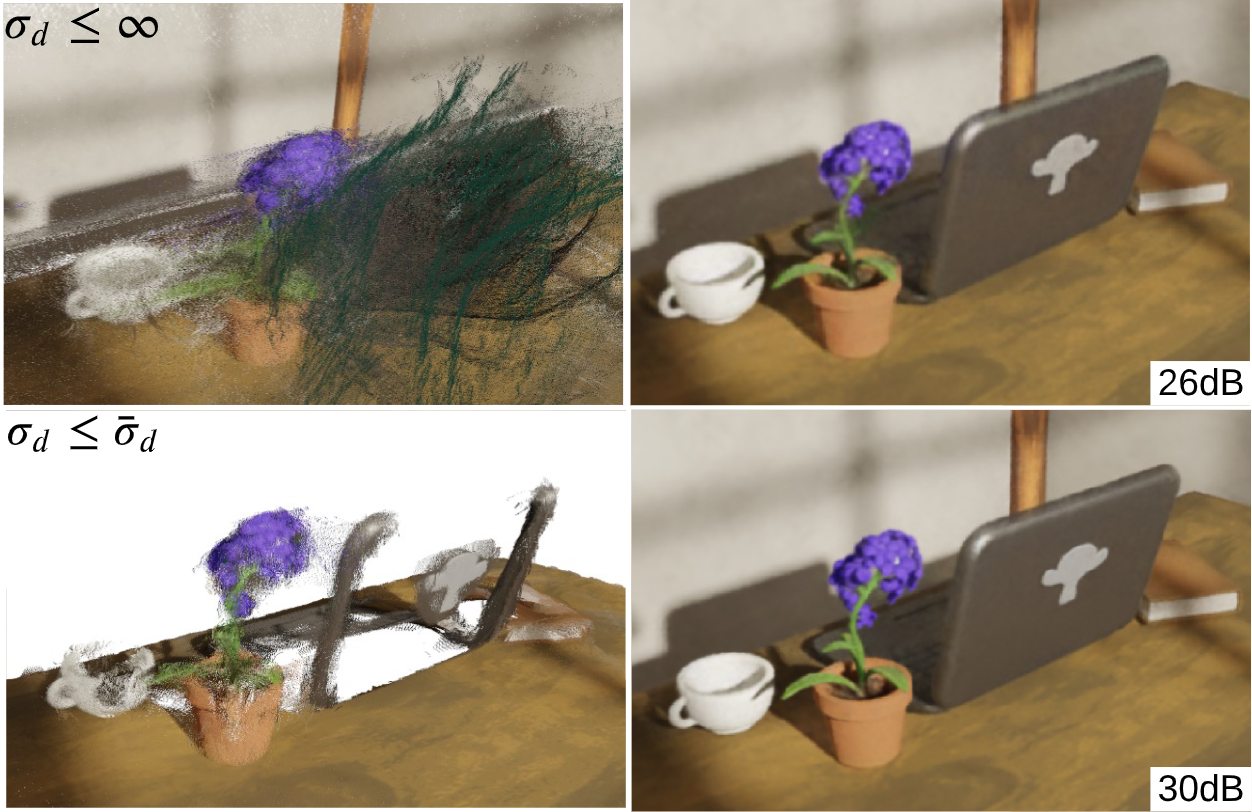}
    \caption{(Top-Left) Raw pointcloud estimated by the tracking module, (Bottom-Left)
    Pointcloud after thresholding the depth uncertainty ($\sigma_d\leq\maxCov$) for visualization.
    (Right Column) Radiance field reconstructions after $120$s of optimization,
     with and without depth weighting (top-right and bottom-right respectively).
    Room scene in Cube-Diorama dataset~\cite{abou2022implicit}.
    }
    \vspace{-1em}
    \label{fig:pcls_with_result}
\end{figure}

In particular, in \cref{fig:pcls_with_result},
we show that if the noisy depths (top-left image) are used as priors without weighting them by their covariance,
the convergence is slower and biased.
PSNR and L1-depth are worst by $4$dB and $7$cm, respectively, after $120$s (top-right image).
Instead, our approach is resilient to these depths,
if we weight them by their marginal covariance (bottom-right image).

\begin{figure}[h!]
    \centering
    \includesvg[width=1.0\columnwidth]{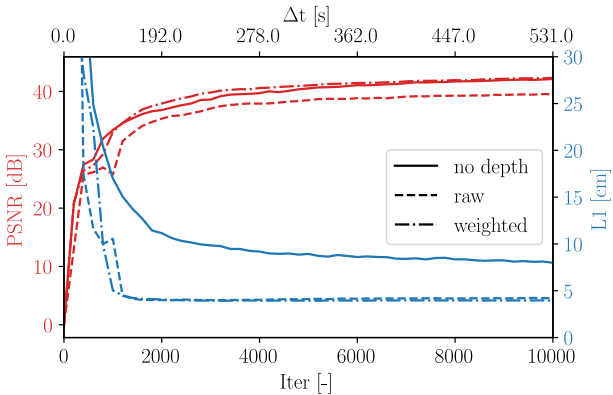}
    \caption{Convergence plot comparing the PSNR and L1 depth metrics over time ($\Delta$t[s]) and iteration (Iter[-]) when not using depths (`no depth'),
     when using the raw depths without weighting (`raw'), and using our approach with weighted depth (`weighted').
    Room scene in Cube-Diorama dataset~\cite{abou2022implicit}.
    }
    \vspace{-1em}
    \label{fig:convergence_plot}
\end{figure}

\cref{fig:convergence_plot}'s convergence plot shows that if we do not use depths,
and only leverage the poses from our tracking module,
the photometric convergence is already good, showing that the pose estimates are sufficiently accurate for convergence.
Nonetheless, the radiance field depth estimates are inaccurate ($7.8$cm L1 depth error after $500$s).
If we use our dense depths (`raw' in \cref{fig:convergence_plot}), the depth error reduces by almost half, to $4.1$cm.
Unfortunately, using depth without uncertainty weighting leads to worst PSNR ($3$dB) than when not using depths,
as it may bias the geometry of the scene.
Finally, weighting the tracking's depth estimates achieves the best PSNR and L1 depth metrics:
it achieves similar PSNR than when not using depths, while having the same L1 error than when using depths.

\poseLoss{\textbf{Uncertainty-weighted Pose Priors.}
While previous approaches have not used poses as priors for mapping,
but rather use pose estimates as initial guesses or as ground-truth poses, we 
use uncertainty-weighted pose priors for mapping.

\TODO{Ablate our pose priors, Figure with pose, wo pose (noisy), with depth, wo depth (none)}
}

\subsection{Real-Time Performance}

Our pipeline is capable of running at $12$ frames per second, with images of $640 \times 480$ resolution.
The tracking thread runs at $15$ frames per second on average, creating around $10$ keyframes per second, depending on the amount of motion.
We add a keyframe every time the mean optical flow is larger than $\meanFlow$ pixels.
The mapping thread runs at $10$ frames per second on average.
Overall, our pipeline is able to reconstruct the scene in real-time at $10$ frames per second,
by parallelizing camera tracking and radiance field reconstruction, and by using custom CUDA kernels.

\section{Limitations}

Our approach requires $\sim$11Gb of GPU memory to operate, given that we use 
dense correlation volumes between pairs of images for tracking and that we use hierarchical volumetric grids for mapping.
The resulting memory requirements can be prohibitively large for robotics applications with low compute power, such as for drones.

Nevertheless, these memory requirements can be alleviated in two ways.
On the one hand, correlation volumes can be computed on the fly, as noted in \cite{teed2020raft}, reducing the need to store dense correlation volumes.
On the other hand, the volumetric information can be streamed to the CPU for inactive regions,
only loading to GPU a sliding window around the region of interest, similarly to \cite{Whelan12rgbd}.

\section{Conclusion}

We show that dense monocular SLAM provides the ideal information for building a \nerf representation of a scene from 
a casually taken monocular video.
The estimated poses and depth-maps from dense SLAM,
weighted by their marginal covariance estimates,
provide the ideal source of information to
optimize a hierarchical hash-based volumetric neural radiance field.
With our approach, users can generate a photometrically and geometrically accurate reconstruction of the scene in real-time.

Future work can leverage our approach to extend the definition of metric-semantic SLAM~\cite{Rosinol21ijrr-Kimera},
which typically only considers geometric and semantic properties, by building representations that are also photometrically accurate.
Beyond metric-semantic SLAM, our approach can be used as the mapping engine for high-level scene understanding,
such as for building 3D Dynamic Scene Graphs~\cite{Armeni16cvpr-3DsemanticParsing,Rosinol20rss-dynamicSceneGraphs,ost2021neural}.

\section*{Acknowledgments}
  This work is partially funded by `la Caixa' Foundation (ID 100010434), LCF/BQ/AA18/11680088 (A. Rosinol), `Rafael Del Pino' Foundation (A. Rosinol), 
  ARL DCIST CRA W911NF-17-2-0181, and ONR MURI grant N00014-19-1-2571.
  We thank Bernardo Aceituno and Yen-Chen Lin for helpful discussions.

{\small
\bibliographystyle{ieee_fullname}
\bibliography{references/refs,references/myRefs}
}

\end{document}